\documentclass[lettersize,journal]{IEEEtran}
\usepackage{amsmath,amsfonts}
\usepackage{algorithmic}
\usepackage{algorithm}
\usepackage{array}
\usepackage[caption=false,font=normalsize,labelfont=sf,textfont=sf]{subfig}
\usepackage{textcomp}
\usepackage{stfloats}
\usepackage{url}
\usepackage{multirow}
\usepackage{verbatim}
\usepackage{graphicx}
\usepackage{cite}
\usepackage{hyperref}
\hypersetup{
	colorlinks=true,
	linkcolor=blue,
	filecolor=blue,      
	urlcolor=black,
	citecolor=blue,
    }

\usepackage{amsmath}
\usepackage{amssymb}
\usepackage{pifont}
\usepackage{threeparttable}

\begin{document}

\title{Revisiting 360 Depth Estimation with PanoGabor: A New Fusion Perspective}

\author{Zhijie~Shen, Chunyu~Lin, \IEEEmembership{Member,~IEEE}, Lang Nie, \IEEEmembership{Member,~IEEE}, Kang~Liao, \IEEEmembership{Member,~IEEE}, \\Weisi~Lin, \IEEEmembership{Fellow,~IEEE}, and Yao~Zhao, \IEEEmembership{Fellow,~IEEE}
\thanks{Zhijie Shen, Chunyu Lin, and Yao Zhao are with the Institute of Information
Science, Beijing Jiaotong University, Beijing 100044, China, and also with
Visual Intelligence +X International Cooperation Joint Laboratory of MOE,
Beijing 100044, China (e-mail: zhjshen@bjtu.edu.cn, cylin@bjtu.edu.cn, yzhao@bjtu.edu.cn).}
\thanks{Lang Nie is with the College of Artificial Intelligence,
Chongqing University of Posts and Telecommunications, Chongqing 400065,
China (e-mail: nielang@bjtu.edu.cn).}
\thanks{Kang Liao and Weisi Lin are with the College of Computing and Data Science, Nanyang Technological University, Singapore (e-mail: kang.liao@ntu.edu.sg, wslin@ntu.edu.sg).}

\thanks{This work was supported by the National Natural Science Foundation of
China (NSFC) under Grant (U2441242,62172032) (Corresponding author: Chunyu Lin)}}

\markboth{Journal of \LaTeX\ Class Files,~Vol.~14, No.~8, August~2021}%
{Shell \MakeLowercase{\textit{et al.}}: A Sample Article Using IEEEtran.cls for IEEE Journals}


\maketitle

\begin{abstract}
Depth estimation from a monocular 360 image is important to the perception of the entire 3D environment. However, the inherent distortion and large field of view (FoV) in 360 images pose great challenges for this task. To this end, existing mainstream solutions typically introduce additional perspective-based 360 representations (\textit{e.g.}, Cubemap) to achieve effective feature extraction. Nevertheless, regardless of the introduced representations, they eventually need to be unified into the equirectangular projection (ERP) format for the subsequent depth estimation, which inevitably reintroduces additional distortions. In this work, we propose an oriented-distortion-aware Gabor Fusion framework (PGFuse) to address the above challenges. First, we introduce Gabor filters that analyze texture in the frequency domain, extending the receptive fields and enhancing depth cues. To address the reintroduced distortions, we design a latitude-aware distortion representation to generate customized, distortion-aware Gabor filters (PanoGabor filters). Furthermore, we design a channel-wise and spatial-wise unidirectional fusion module (CS-UFM) that integrates the proposed PanoGabor filters to unify other representations into the ERP format, delivering effective and distortion-aware features. Considering the orientation sensitivity of the Gabor transform, we further introduce a spherical gradient constraint to stabilize this sensitivity. Experimental results on three popular indoor 360 benchmarks demonstrate the superiority of the proposed PGFuse to existing state-of-the-art solutions. Code and models will be available at \url{https://github.com/zhijieshen-bjtu/PGFuse}. 
\end{abstract}


\begin{figure}[t]
  \centering
  \includegraphics[width=0.47\textwidth]{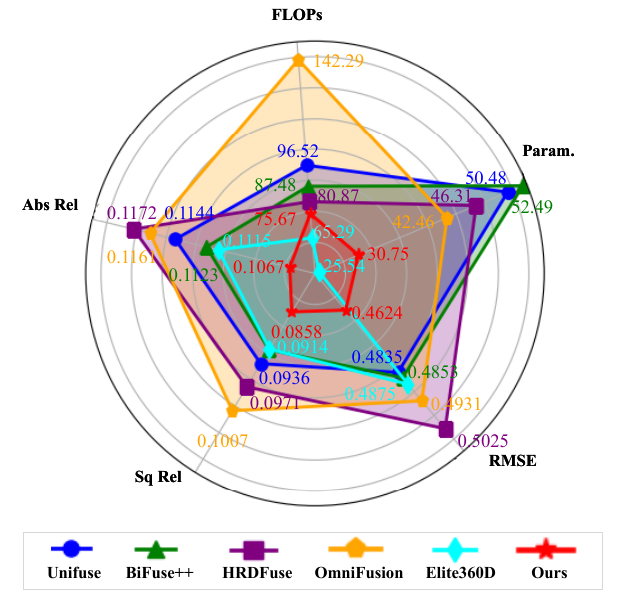} 
  \caption{Comparison of our proposed framework (shown in red) against other existing methods. Our method consistently demonstrates lower values across all error metrics. The compact red area highlights that our approach maintains competitive scores in computational efficiency (FLOPs), demonstrating a well-balanced trade-off between performance and computational cost.} 
  \label{fig:door}
\end{figure}

\section{Introduction}
\IEEEPARstart{3}{60} images provide panoramic information of the surroundings with a single shot, while monocular 360 depth estimation enables the perception of the complete 3D scene information. However, mainstream models are specifically designed for perspective images and cannot address the challenges of distortions and the large FoV posed by 360 images due to fixed sampling positions and limited receptive fields. Consequently, addressing panoramic distortion and extending the receptive field have become primary concerns of current methodologies.

To mitigate the negative effects of distortion, BiFuse \cite{Wang2020BiFuseM3} introduced a dual-projection fusion scheme, which captures distortion-aware features using an additional cubemap representation. However, its bidirectional fusion module imposes significant computational and memory burdens. To address this issue, Unifuse \cite{Jiang2021UniFuseUF} proposes a more efficient unidirectional fusion module based on residual learning to reduce computational demands. Moreover, HRDfuse \cite{Ai2023HRDFuseM3} replaces the cubemap projection with a more advanced Tangent Patch projection format (TP), achieving an effective fusion of TP and ERP features through a spatial feature alignment module. 
On the other hand, some researchers focus on introducing distortion awareness to transformer architectures with large receptive fields \cite{Shen2022PanoFormerPT,Yun2023EGformerEG,Ling2023PanoSwinAP}. Considering the non-negligible computational and memory cost inherent in transformers, Elite360D \cite{Ai_2024_CVPR} proposes a bi-projection bi-attention fusion module that integrates ERP and icosahedron projection (ICOSAP) \cite{Cohen2019GaugeEC,Jiang2019SCUG} features, combining the strengths of fusion-based schemes and transformer-based architectures. However, this type of fusion-based method, whether introducing new representations or proposing new fusion modules, overlooks the fact that all representations need to be integrated into the ERP format for fusion, which inevitably reintroduces distortion. Therefore, \textit{we argue that explicitly addressing the troublesome panoramic distortion in the fusion process can benefit the subsequent depth estimation.} 

To this end, we present an oriented-distortion-aware Gabor Fusion framework (termed PGFuse), an oriented panoramic Gabor-based fusion framework for monocular omnidirectional depth estimation. PGFuse can address reintroduced distortions when integrating other representations into the ERP format. Specifically, we explore the Gabor wavelets for panoramic depth estimation, aiming to leverage the advantages of a human vision-inspired model. Recent research has focused on extracting effective features based on human visual principles \cite{shen2006review,meshgini2012face}. The Gabor wavelet kernel, resembling the two-dimensional receptive field of human vision, is highly effective at capturing key visual features such as spatial positioning, directional selectivity, and spatial frequency. Unlike other methods that extend the receptive field, Gabor filters are particularly effective in panoramic depth estimation due to their ability to capture multi-scale spatial frequencies and directional information simultaneously. This enables them to handle better the large, varying depth ranges and complex textures typically encountered in panoramic scenes, improving depth estimation accuracy. 
Besides, the tunable parameters of Gabor filters enable the creation of task-specific filters without additional parameters. However, the inherent distortion in 360 images poses a significant challenge for Gabor filter applications. To address this challenge, we design Gabor filters using a latitude-aware distortion representation method, 
which adjusts the parameters of Gabor filters to achieve distortion awareness. These finely tuned Gabor filters, termed PanoGabor filters, are then incorporated as a convolutional layer. PanoGabor is embedded into a specially designed Channel-wise and Spatial-wise Unidirectional Fusion Module (CS-UFM) to mitigate the reintroduced distortions. More importantly, the parameters of PanoGabor can be optimized via dataset-driven training, ensuring seamless integration. Furthermore, we introduce a spherical gradient technique to stabilize the orientation sensitivity of the Gabor filters. Leveraging Gabor-based distortion-aware design and a perspective-based feature extraction strategy, PGfuse achieves the best-balanced trade-off between performance and computational cost, as shown in Fig. \ref{fig:door}.

Extensive experiments evaluated on three popular indoor datasets Stanford2D-3D \cite{Armeni2017Joint2D}, Matterport3D \cite{Chang2017Matterport3DLF}, and Structured3D \cite{Zheng2019Structured3DAL} demonstrate that our proposed scheme outperforms the state-of-the-art approaches. Our contributions can be summarized as follows:
\begin{itemize}
    \item We introduce PGFuse, a panoramic Gabor-based fusion framework that effectively addresses the reintroduced distortion problems that are often overlooked by previous methods.
    \item To address the challenges of large FoV and panoramic distortion of 360 images, we design panorama-specific Gabor filters to analyze panoramic texture in the frequency domain, thereby extending the receptive fields and enhancing depth cues. 
    \item We design a channel-wise and spatial-wise unidirectional fusion module for the integration of other representations into ERP format, delivering effective and distortion-aware features for the subsequent depth map estimation.
\end{itemize}
\begin{figure*}[t]
  \centering
  \includegraphics[width=\textwidth]{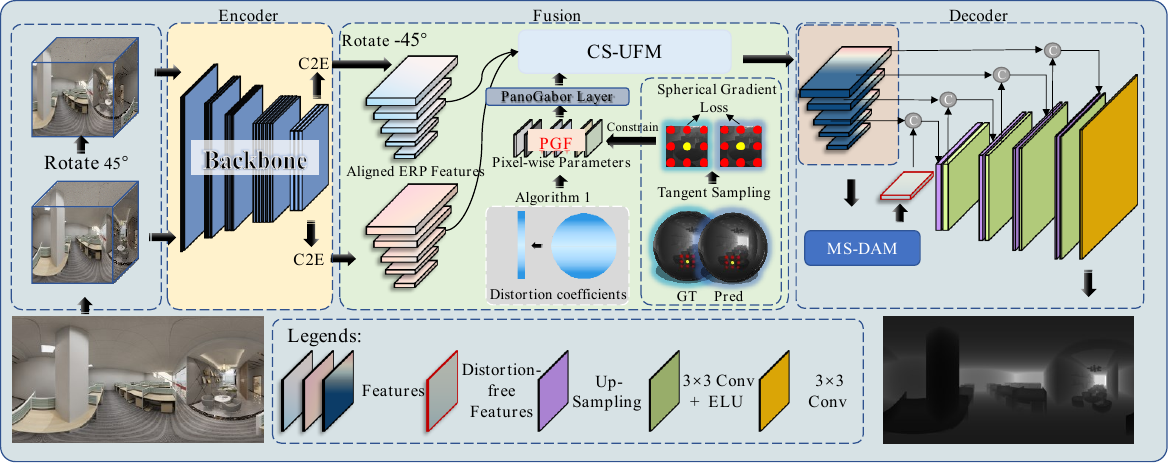} 
  \caption{Overview of the proposed framework. Our framework takes a single panorama as input and outputs the corresponding depth map. This network addresses the reintroduced distortion problems that are often overlooked by previous
  fusion-based methods. Specifically, we introduce a PanoGabor approach to achieve the distortion awareness.} 
  \label{fig:framework}
\end{figure*}

\section{Related work}
\label{gen_inst}
\subsection{Monocular 360 Depth Estimation}
Panoramic distortion leads to notable performance degradation in perspective image-based depth estimation approaches. To this end, Zioulis $et~ al.$~\cite{Zioulis2018OmniDepthDD} highlighted the need to learn within the 360 domain, introducing the 3D60 dataset. Moreover, to mitigate distortion, researchers have adjusted the sampling positions of convolutional kernels for distortion awareness~\cite{jiang2019spherical,xu2021spherical,chen2021distortion}. Some indoor monocular panoramic depth estimation methods~\cite{Cheng2020OmnidirectionalDE,Zhuang2021ACDNetAC,Li2023mathcalA} have developed various distortion-aware techniques to manage panoramic distortions, but most of these techniques are computationally inefficient. Recently, advanced Transformer-based architectures~\cite{Liu2021SwinTH} have broadened the receptive field, enabling models to extract features from 360 images. However, the traditional self-attention mechanism struggles to handle the complex distortion distribution in 360 images~\cite{Shen2022PanoFormerPT}. Enhanced distortion-aware transformer architectures~\cite{Shen2022PanoFormerPT,Yun2023EGformerEG,Ling2023PanoSwinAP,Li2022OmniFusion3M,zhuang2023spdet,Shen2023Disentangling}, while effective, often come with high computational and memory costs. To simplify models, some approaches~\cite{Sun2020HoHoNet3I,Sun2019HorizonNetLR,Pintore2021SliceNetDD,shen2021distortion,cheng2018cube} advocate for estimating indoor depth from vertically compressed one-dimensional sequences, which leverages gravity alignment properties. However, this compression often fails to capture essential structural geometric information. To achieve a balanced trade-off between performance and efficiency, the mainstream approaches\cite{Yoon2021SphereSRI,Ai_2023_CVPR,Ai_2024_CVPR} still prefer to develop a more efficient fusion-based framework. However, these fusion methods neglect the reintroduced distortion during the fusion process. Besides, although many of these approaches claim state-of-the-art performance, comparisons are often complicated by variations in datasets, preprocessing methods, and evaluation strategies, making fair comparisons challenging. To this end, Elite360D~\cite{Ai_2024_CVPR} provides a solid baseline. Therefore, in this paper, we follow the Elite360D method~\cite{Ai_2024_CVPR} to conduct the experiments.  Different from prior fusion frameworks such as UniFuse \cite{jiang2021unifuse}, which rely on ERP–cubemap pairs and simple concatenation strategies, our method adopts dual cubemaps with a single shared-weight encoder and introduces a distortion-aware PanoGabor fusion layer to explicitly correct reintroduced ERP distortions. 
To our knowledge, this is the first framework that explicitly addresses the problem of reintroduced distortions in fusion, thereby defining a novel paradigm for panoramic depth estimation beyond existing dual-projection fusion-based architectures.

\subsection{Gabor Filters in Image Processing}
The Gabor filter is widely used in texture analysis, object detection, and face recognition due to its robustness to varying angles and scales. Li $et~ al.$~\cite{li2021learning} suggested enhancing face recognition by learning the Gabor wavelet's covariance matrix. In fabric defect detection, Chen $et~ al.$~\cite{chen2022improved} introduced the Genetic Algorithm Gabor Faster R-CNN (Faster GG R-CNN), embedding Gabor kernels to reduce texture interference. Andrey $et~ al.$~\cite{alekseev2019gabornet} proposed GaborNet, which uses Gabor filters to improve convergence and reduce the complexity of training in image recognition. To extend the receptive fields and enhance the geometric depth cues, we introduce Gabor filters to analyze texture in the frequency domain. Considering the special characteristics of 360 images, we propose a distortion-aware Gabor-based approach to address the challenging distortion issues.

\section{Methodology}
\label{headings}
 In this section, we introduce PGFuse, whose overall pipeline is illustrated in Fig.~\ref{fig:framework}. The proposed framework is designed as a two-stage distortion-optimization process, in which cubemap-based feature extraction and PanoGabor-based fusion are deployed in complementary domains to address distinct distortion sources. Specifically, equirectangular projection (ERP) suffers from severe non-uniform geometric distortion, particularly near the poles, which greatly hinders the effectiveness of convolutional feature extraction. While PanoGabor is effective in refining distortion-aware frequency responses, it cannot fundamentally remove the intrinsic geometric deformation of ERP. To mitigate this issue, we first project the sphere into the cubemap domain, which decomposes the panorama into locally tangential planes with significantly reduced geometric distortion. This design also allows us to take advantage of perspective-pretrained backbones, thereby improving both training efficiency and feature quality. To further alleviate pixel-missing and discontinuous boundary problems, we adopt a dual-cube strategy, in which two cubemaps—one in the original orientation and one rotated by 45$^\circ$—are jointly exploited~\cite{shen2021distortion}. These features are spatially aligned, unified into the ERP format, and subsequently processed by our distortion-aware fusion mechanism. 

At this stage, a second type of distortion inevitably arises: projection-induced resampling artifacts and cube-face seams during the unification back into ERP space. To explicitly address this problem, we design PanoGabor filters and embed them into the CS-UFM fusion module. The PanoGabor filters incorporate a latitude-aware distortion representation that approximates the cosine weighting of latitude and modulates frequency-domain responses accordingly. In this way, they enhance depth cues, smooth boundary discontinuities, and ensure global consistency of the ERP representation. Finally, the refined features are fed into a multi-scale distortion-aware module (MS-DAM)~\cite{Shen2023Disentangling} to integrate shallow geometric and deep semantic cues, followed by a hierarchical up-sampling decoder that produces the final depth map. Additional spherical gradient constraints are imposed by projecting the ERP depth map onto tangent patches, where Sobel operators are applied to stabilize orientation sensitivity. The following subsections present the details of each component.

\begin{figure}[t]
  \centering
  \includegraphics[width=0.5\textwidth]{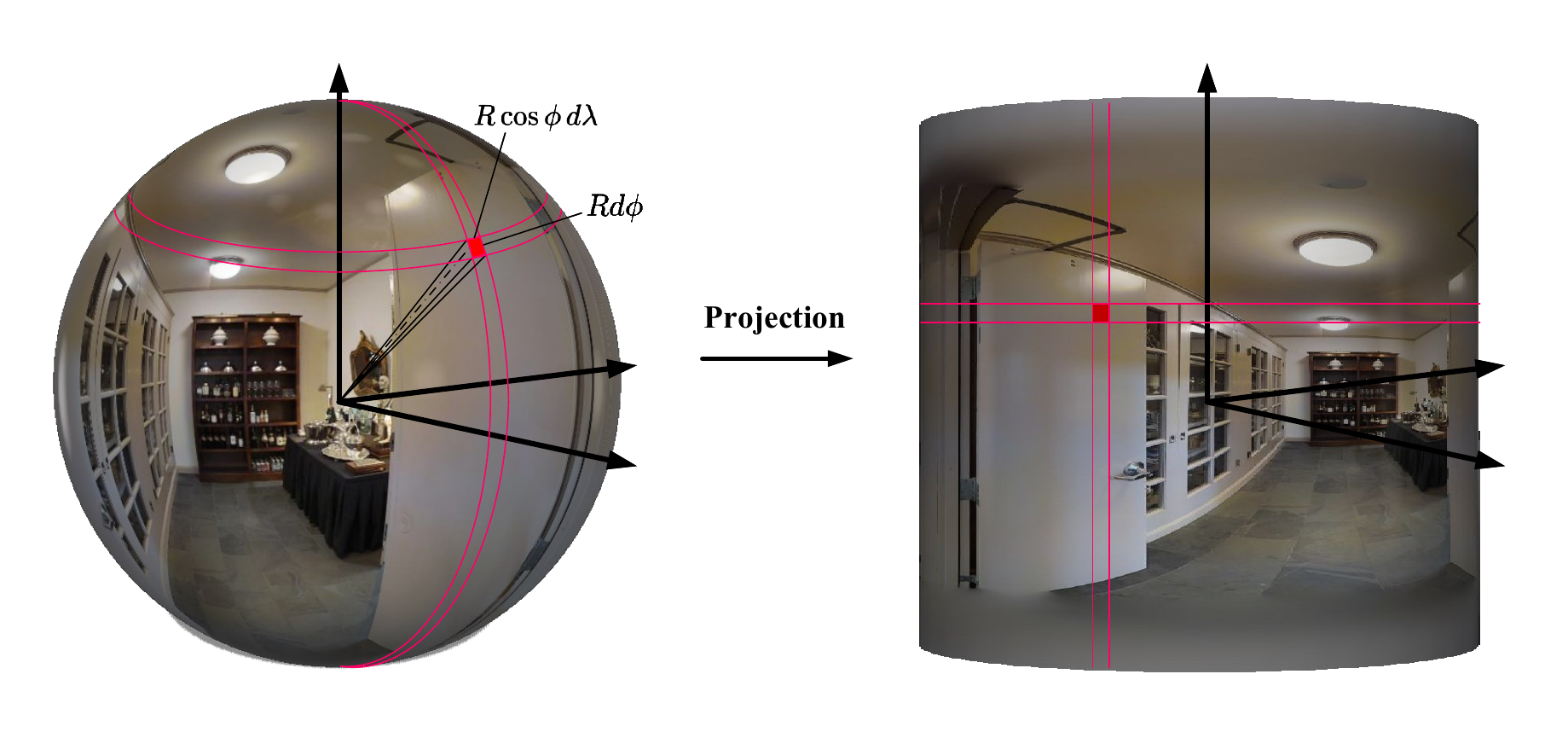} 
  \caption{Illustration of the differential area element on the spherical surface (left) and the ERP domain (right).} 
  \label{fig:diffarea}
\vspace{-10pt}
\end{figure}
\subsection{Preliminary}
\label{sec_pre}
\noindent \textbf{Distortion Distribution Pattern.} In the context of ERP projections, the distortion distribution increases progressively from the equatorial region towards the poles. To mathematically describe this distribution, we introduce a method based on the ratio of differential area elements before and after the ERP projection operation. Given a small region on the surface of a sphere, with spherical coordinates defined by the center of the sphere as the origin, we denote the differentials of latitude \( \phi \) and longitude \( \lambda \) by \( d\phi \) and \( d\lambda \), respectively. 
When the latitude changes from \( \phi \) to \( \phi + d\phi \), the corresponding arc length is \( R d\phi \), where \( R \) is the radius of the sphere. At latitude \( \phi \), the arc length in the longitude direction is \( R \cos(\phi) d\lambda \), since the distance corresponding to a unit change in longitude decreases as latitude increases. Consequently, the differential area element of the region on the spherical surface (Fig. \ref{fig:diffarea}(left)) can be expressed as:

\begin{equation}
dA = R^2 \cos(\phi) \, d\phi \, d\lambda,
\end{equation}
where \( \cos(\phi) \) accounts for the latitude-dependent scaling factor. However, in the ERP domain, the variations in both latitude and longitude are assumed to be uniform due to the stretching effect. As a result, the differential area element in this domain (Fig. \ref{fig:diffarea}(right)) is approximated as:

\begin{equation}
dA' = R d\phi \, R d\lambda = R^2d\phi \,d\lambda.
\end{equation}
The distortion factor \( K(\phi) \) is then defined as the ratio of the differential area element in the ERP projection to the differential area element on the spherical surface:

\begin{equation}
K(\phi) = \frac{dA'}{dA} = \frac{1}{ \cos(\phi)}.
\end{equation}
The distortion factor \( K(\phi) \) ranges from \( 1 \) to \( +\infty \). At the equator, the projection corresponds to an identity mapping, implying that the distortion is zero at this latitude. To more accurately describe the variation in distortion across latitudes, we introduce a bias term, leading to the following expression for the distortion factor:

\begin{equation}
K(\phi) = \frac{1}{\cos(\phi)} - 1.
\end{equation}
This formulation characterizes the distortion as a function of latitude in the ERP projection more precisely. It provides a foundation for further distortion analysis and potential compensation techniques in the context of ERP projections.

\noindent \textbf{Gabor Transform.} When applied to image processing, the Gabor transform can provide useful spatial frequency and orientation analysis of the images. By responding to specific spatial frequencies and orientations, the Gabor filters can effectively identify textures and detect edges, which are critical for depth estimation. The standard Gabor transform formula can be denoted as:
\begin{eqnarray}
\label{eq:real}
g(x,y,f,\theta,\psi,\sigma) = \frac{1}{2\pi\sigma^2} e^{-\frac{x^{'2}+y^{'2}}{2\sigma^2}}\cos(fx^{'}+\psi),
\end{eqnarray}where
\begin{eqnarray}
 x^{'}&=& x\cos\theta + y\sin\theta \\ 
y^{'}&=& -x\sin\theta + y\cos\theta,     
\end{eqnarray}
and $(x, y)$ represents the coordinates of the Gabor kernels (the pixel position in the spatial domain); the kernel size is set to 3 (\textit{i.e.}, $x, y \in \{-1, 0, 1\}$); $f$ indicates the central angular frequency of a sinusoidal plane wave, $\theta$ represents the orientation of the Gabor filter, $\psi$ is the phase offset, $\sigma$ represents the sharpness of the Gaussian function along both $x$ and $y$ directions.

\begin{algorithm}[tb]
\caption{Distortion-Aware PanoGabor Filters}
\label{alg:algorithm}
\textbf{Input:}\\
$C_{in}$: input channel numbers;\\
$C_{out}$: output channel numbers; \\
$K(\phi_C) \in \mathbb{R}^{C_{in}}$: distortion coefficients;\\  
\textbf{Output:} PanoGabor Filters
\begin{algorithmic}[1]
\STATE Define 8 orientations: $\{\theta_i \in \mathbb{R}^{\frac{C_{in}}{8}\times C_{out}} = \frac{\pi \cdot i}{8} \}_{i=0}^7$
\STATE Divide $C_{in}$ into 8 groups:\\
\FOR{$i^{th}$ group}
\STATE apply $\theta_i$
\ENDFOR
\STATE $\theta \in \mathbb{R}^{C_{in}\times C_{out}} \leftarrow$ concatenate
\STATE Calculate Gabor parameters\footnotemark:\\
$\psi \in \mathbb{R}^{C_{in}\times C_{out}} \leftarrow \pi \epsilon$; \\
$f \in \mathbb{R}^{C_{in}\times C_{out}} \leftarrow (\frac{\pi}{2}) \cdot (\sqrt{2})^{\epsilon}$;\\ 
$\sigma \in \mathbb{R}^{C_{in}\times C_{out}} \leftarrow \frac{\pi}{f + 0.1}$
\STATE  
$f \leftarrow f \cdot (1 + K(\phi_C))$;
\STATE $(y, x) \leftarrow$ meshgrid$((-1, 0, 1), (-1, 0, 1))$; $x, y \in \mathbb{R}^{3 \times 3}$
\STATE Rotate $x$ and $y$:\\
    $x_{rot} \leftarrow x \cdot \cos(\theta) + y \cdot \sin(\theta)$;\\
    $y_{rot} \leftarrow -x \cdot \sin(\theta) + y \cdot \cos(\theta)$
\STATE Calculate Gabor filter $g$:\\
    $g \leftarrow \exp\left(-0.5 \cdot \left(\frac{x_{rot}^2 + y_{rot}^2}{\sigma^2}\right)\right) \cdot \cos(f \cdot x_{rot} + \psi)$;\\
    $g \leftarrow \frac{g}{2 \cdot \pi \cdot \sigma^2}$;\
\STATE Reshape $g \in \mathbb{R}^{C_{in}\times C_{out}\times 3 \times 3}$
\RETURN $g$
\end{algorithmic}
\end{algorithm}
\footnotetext{The parameters are set up according to \cite{meshgini2012face}, $\epsilon$ is generated randomly.}

\subsection{Distortion-aware  PanoGabor Filters (PGF)}
\label{sec_3_2}
We aim to utilize the advanced texture analysis capabilities and inherent frequency domain transformation properties of Gabor filters to address the challenges posed by the large FoV in 360 images. However, due to the severe distortion, directly incorporating the standard Gabor filters into the learning frameworks for monocular 360 depth estimation would lead to unsatisfactory prediction. In contrast, the learnable Gabor filters can dynamically adapt to the distortions but deliver suboptimal performance because the distortion distribution pattern is hard to model in a data-driven way. Inspired by the spherical projection process, we propose calculating the distortion coefficients for distortion-aware Gabor filter generation. Here, we propose the latitude-aware distortion representation method, which approximates the distortion coefficients as a function of the latitude.  

The approximation derives from the equirectangular projection and the geometric properties of a sphere. Due to the spherical representation, the distance covered by a degree of longitude decreases from the equator towards the two poles. For latitudes \(\phi \in (-\frac{\pi}{2}, \frac{\pi}{2})\), the distortion coefficient $d_c$  can be approximated by the reciprocal of the cosine of the latitude (the proof can be found in Sec. \ref{sec_pre}):
\begin{eqnarray}
\label{eq:linear}
K(\phi) = \frac{1}{\cos(\phi)} - 1,
\end{eqnarray}
where the range of $d_c$ is $[0, +\infty)$. The scale factor mirrors the increasing distortion in equirectangular projections towards the poles. 
Subsequently, as detailed in Algorithm \ref{alg:algorithm}, we employ the defined distortion coefficient to control the generation of distortion-aware PanoGabor filters, thereby creating a bank of PanoGabor filters.

\begin{figure}[t]
  \centering
  \includegraphics[width=0.3\textwidth]{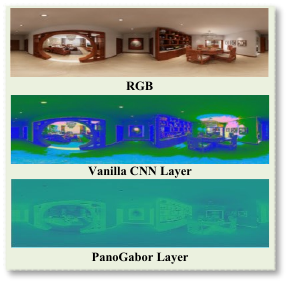} 
  \caption{Visualization of the features extracted through the vanilla CNN layer and our proposed PanoGabor layer. Compared to the vanilla CNN layer, our PanoGabor layer is more robust regarding surface textures and complex illumination variations, which is crucial for accurate indoor depth estimation.} 
  \label{fig:pg}
\end{figure}

\noindent\textbf{PanoGabor Convolution Layer.} We further enclose the PanoGabor filters as an independent convolution layer. Given a set of input features \(\mathcal{F} \in \mathbb{R}^{B \times C \times H \times W}\), we initially apply a standard convolutional layer to transform these features into channel-height dimension-matched features, denoted as \(\mathcal{F'} \in \mathbb{R}^{B \times H (C_{in}) \times H \times W}\). Firstly, we calculate the distortion coefficient for each channel. Let \( C = [ 0, 1, \dots, C_{in}-1 ] \) be the latitude grid indices, the latitude value can be calculated as follows:
\begin{eqnarray}
\phi_C = \left\{ (\frac{c_i}{C_{in}} - 0.5)\pi| c_i\in C  \right \}. 
\end{eqnarray}
We then apply Eq. \ref{eq:linear} to calculate the distortion coefficients $K(\phi_C) \in \mathbb{R}^{C_{in}}$ for all channels. Subsequently, we use Algorithm 1 to generate a set of PanoGabor filters. These filters are then employed to perform the convolution operation, compensating for the geometric distortions in spherical features. By making the PanoGabor layer learnable, the filters become distortion-aware and data-driven. This allows the PanoGabor layer to integrate seamlessly with the subsequent CS-UFM module, facilitating joint optimization. As illustrated in Fig. \ref{fig:pg}, compared to the vanilla CNN layer, our PanoGabor layer demonstrates enhanced robustness to interference from surface textures and complex illumination variations in indoor scenes, which is crucial for accurate indoor depth estimation. Furthermore, our method remains unaffected by distortions, thereby effectively extracting ERP features with explicit geometric structures from 360 images.
\begin{figure}[t]
  \centering
  \includegraphics[width=0.5\textwidth]{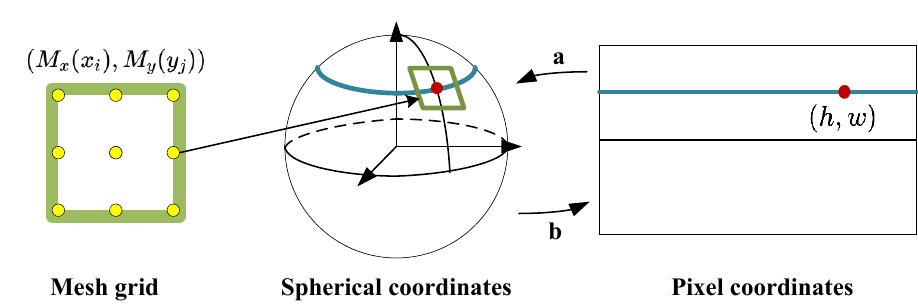} 
  \caption{Illustration of the spherical tangent sampling. A representation that the pixels are projected from the pixel coordinates to the spherical coordinates to sample tangent points. b indicates that the calculated sampling points are projected back to the pixel coordinates.} 
  \label{fig:sgd}
\end{figure}
\subsection{Spherical Gradient Constraint}
\label{sec_3_3}
We have introduced our PanoGabor approach, which extends the original Gabor filters to be learnable and panorama-specific to achieve the dynamic perception of panoramic features. However, because of the sensitivity to the orientations, PanoGabor does not work well when the original gradient constraints are naively applied. Therefore, we perform a spherical gradient constraint for PanoGabor to assist in the dynamic parameter adjustment. Specifically, we calculate gradients on a pixel-level tangent plane, following the methodology outlined in previous works \cite{ Eder2019TangentIF, Yuan2021360OF}. 

\textbf{Create sampling mesh grid.} Given a depth map \( D \in \mathbb{R}^{H \times W} \), we first define the sampling grid. For each pixel in the depth map, we sample a 3×3 tangent patch. Let \( X = [-1, 0, 1] \) and \( Y = [-1, 0, 1] \) represent the sampling kernels for the $x$- and $y$-directions, respectively. We define the 2D grid mesh $M$ (Fig. \ref{fig:sgd}left) as:

\begin{equation}
M = \left\{ (M_x(x_i), M_y(y_j)) \mid x_i \in X, y_j \in Y \right\},
\end{equation}
where \( M_x(x_i) \) and \( M_y(y_j) \) represent the corresponding coordinates in the tangent plane for the sampling positions:

\begin{equation}
\begin{aligned}
M_x(x_i) &= \tan(x_i \cdot \Delta \lambda), \\
M_y(y_j) &= \frac{\tan(y_j \cdot \Delta \phi)}{\cos(y_j \cdot \Delta \lambda)}.
\end{aligned}
\end{equation}
Here, \( \Delta \lambda = \frac{2\pi}{W} \) and \( \Delta \phi = \frac{\pi}{H} \) denote the sampling step sizes in the longitude \(\lambda\) and latitude \(\phi\) directions, respectively. This ensures that the sampling grid appropriately covers the spherical surface based on the pixel indices.

\textbf{Pixel coordinates to spherical coordinates (Fig. \ref{fig:sgd}a).} For each pixel located at row $h$ and column $w$ of the depth map, we compute the corresponding latitude and longitude values in spherical coordinates. The latitude $\phi_h$ for row $h$ is calculated as:
\begin{equation}
\phi_h = \left( \frac{h}{H} - 0.5 \right) \pi,
\end{equation}
and the longitude $\lambda_w$ for column $w$ is given by:

\begin{equation}
\lambda_w = \left( \frac{w}{W} - 0.5 \right) 2\pi.
\end{equation}
These latitude and longitude values correspond to the pixel's position on the spherical surface, with the normalization accounting for the pixel dimensions in the depth map.

\textbf{Calculate radial distance and angular distortion.} For each sampled kernel point at position $(x_i, y_j)$, we compute the radial distance $\rho$, which is the Euclidean distance between the corresponding coordinates in the tangent plane:

\begin{equation}
\rho(x_i, y_j) = \sqrt{M_x(x_i)^2 + M_y(y_j)^2}.
\end{equation}
The radial distance is then used to calculate the angular distortion \( \nu \), which accounts for the curvature of the spherical surface:

\begin{equation}
\nu(x_i, y_j) = \arctan(\rho(x_i, y_j)).
\end{equation}
This angular distortion is crucial for compensating for the non-uniformity introduced by the spherical geometry.

\textbf{Latitude and longitude of tangent grid points.} Using the radial distance $\rho(x_i, y_j)$ and angular distortion $\nu(x_i,y_j)$, we compute the latitude and longitude for each sampling point in the tangent patch. The latitude \( \phi(h, x_i, y_j) \) for the pixel $(h, w)$ and the kernel position $(x_i,y_j)$ is computed as:
\begin{equation}
\begin{aligned}
\phi(h, x_i, y_j) = \arcsin \Big( \cos(\nu(x_i, y_j)) \sin(\phi_h) \\
+ \frac{M_y(y_j) \sin(\nu(x_i, y_j)) \cos(\phi_h)}{\rho(x_i, y_j)} \Big).
\end{aligned}
\end{equation}
Similarly, the longitude \( \lambda(h, x_i, y_j) \) is calculated as:

\begin{equation}
\lambda(h, x_i, y_j) = \lambda_1(h, x_i, y_j) + \lambda_w,
\end{equation}
where $\lambda_1$ is given by:

\begin{equation}
\lambda_1(h, x_i, y_j) = \arctan \left( \frac{M_x(x_i) \sin(\nu(x_i, y_j))}{\Delta} \right),
\end{equation}
and $\Delta$ is computed as:

\begin{equation}
\begin{aligned}
\Delta = \rho(x_i, y_j) \cos(\phi_h) \cos(\nu(x_i, y_j)) \\
- M_y(y_j) \sin(\phi_h) \sin(\nu(x_i, y_j)).
\end{aligned}
\end{equation}

\textbf{Projection to pixel coordinates.} Finally, we project the calculated latitude and longitude values back onto the image pixel coordinates (Fig. \ref{fig:sgd}b). The new row and column positions $h'$ and $w'$ are given by:

\begin{equation}
\begin{aligned}
h' &= \frac{\phi(h, x_i, y_j) H}{\pi} + 0.5, \\
w' &= \left( \frac{\lambda(h, x_i, y_j)}{2\pi} + 0.5 \right) W \mod W, 
\end{aligned}
\end{equation}
where the MOD operation appropriately handles the seamless nature of the left and right boundaries in panoramic images, ensuring that when the sampling range exceeds the boundaries, it is wrapped around to the opposite side. With these projected positions, we sample the corresponding depth values from the depth map. The function \( f_s(*) \) formalizes this sampling process, which allows us to efficiently compute the gradients on the spherical surface.

\textbf{Gradient calculation.} The gradients are then computed by applying the Sobel operator in the $x$- and $y$-directions. The horizontal and vertical gradients $G_x(D)$ and $G_y(D)$ are given by:
\begin{equation}
\begin{aligned}
G_x(D) &= \frac{1}{HW} \sum_{i=0}^{HW-1} f_s(D_i) \otimes S_x, \\
G_y(D) &= \frac{1}{HW} \sum_{i=0}^{HW-1} f_s(D_i) \otimes S_y,
\end{aligned}
\end{equation}
where \( S_x \) and \( S_y \) represent the horizontal and vertical components of the Sobel operator, respectively. These gradients are computed over the entire depth map $D$, where $HW$ is the total number of pixels. The use of the Sobel operator ensures that the gradients capture both the local spatial variations and the curvature effects on the spherical surface.
\begin{figure}[t]
  \centering
  \includegraphics[width=0.5\textwidth]{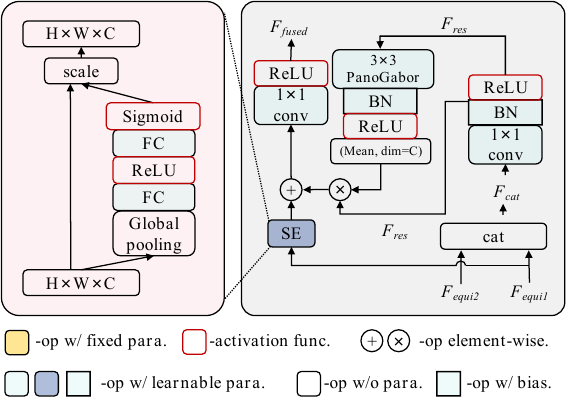} 
  \caption{The proposed CS-UFM module.} 
  \label{fig:csufm}
\end{figure}
\subsection{Channel-wise and Spatial-wise Unidirectional Fusion Module (CS-UFM)}
\label{sec_3_4}
The fusion module is designed to integrate two sets of ERP features, addressing reintroduced distortion, inconsistencies in cubemap boundaries, and pixel loss issues. This module (as illustrated in Fig. \ref{fig:csufm}) first concatenates the two groups of input ERP features and uses a residual block to achieve an initial refinement. In particular, the block comprises a 1x1 convolution that compresses the channels to match the features' height, facilitating the application of the PanoGabor layer. Following the 1×1 convolution layer, a 3×3 PanoGabor layer is used to process the features, aiming for distortion awareness and depth cues enhancement.

Subsequently, the enhanced features are multiplied by the mean values of the residual features on a spatial-wise basis to ensure a distortion-aware feature combination. To further refine the feature representation, a Squeeze-and-Excitation (SE) layer is applied, adaptively recalibrating channel-wise responses to emphasize important features and suppress redundant ones. The final stage involves a 1x1 convolution that produces the output feature map, effectively integrating the strengths of spatial and channel-wise processing.

This design enables the fusion module to enhance the analysis and integration of ERP features, making it particularly effective for subsequent depth estimation. The incorporation of PanoGabor filters and SE mechanisms provides a robust approach to enhance the module's representational capacity, both channel-wise and spatial-wise.   

\begin{table*}[t]
\centering
\begin{threeparttable}[b]
\caption{Quantitative comparison with the SOTA methods, including EGFormer~\cite{Yun2023EGformerEG}, PanoFormer~\cite{Shen2022PanoFormerPT}, BiFuse~\cite{Wang2020BiFuseM3}, BiFuse++~\cite{Wang2022BiFuseSA}, UniFuse~\cite{Jiang2021UniFuseUF}, OmniFusion~\cite{Li2022OmniFusion3M}, HRDFuse~\cite{Ai_2023_CVPR}, and Elite360D~\cite{Ai_2024_CVPR}.
}
    \begin{tabular}{c c c c c c c c c c}
    \hline Datasets & Backbone & Method & Pub'Year  $(\bar{G})$  & Abs Rel  $\downarrow$  & Sq Rel  $\downarrow$  &  $\mathrm{RMSE} \downarrow$  &  $\delta_{1}(\%) \uparrow$  &  $\delta_{2}(\%) \uparrow$  &  $\delta_{3}(\%) \uparrow$  \\
    \hline 
    &Transformer & EGFormer &  ICCV'23  & 0.1473 &  0.1517  & 0.6025 &  81.58  & 93.90 & 97.35 \\
    & & PanoFormer &  ECCV'22   & 0.1051 & 0.0966 & 0.4929 & 89.08 & 96.23 & 98.31 \\
    \cline{2-10} & & BiFuse & CVPR'20    & 0.1126 &  0.0992  & 0.5027 &   88.00  & 96.13 & 98.47 \\
     & & BiFuse++ & TPAMI'22  & 0.1123 &   0.0915  &  0.4853  & 88.12 & 96.56 & 98.69 \\
     M3D& & UniFuse & RAL'21  & 0.1144 & 0.0936 & 0.4835 & 87.85 & 96.59 & 98.73 \\
     & ResNet-34& OmniFusion & CVPR' 22  & 0.1161 & 0.1007 & 0.4931 & 87.72 & 96.15 & 98.44 \\
     & & HRDFuse  & CVPR'23  & 0.1172 & 0.0971 & 0.5025 & 86.74 & 96.17 & 98.49 \\
     & & Elite360D & CVPR'24  & 0.1115 &  0.0914  & 0.4875 & 88.15 & 96.46 & \textbf{98.74} \\
     \cline{3-10}& & Ours &- & \textbf{0.1067} & \textbf{0.0858} & \textbf{0.4624} & \textbf{89.06} & \textbf{96.75} & 98.72 \\
    \hline & Transformer& EGFormer & ICCV'23 & 0.1528 & 0.1408 & 0.4974 & 81.85 & 93.38 & 97.36 \\
     & & PanoFormer &  ECCV'22  & 0.1122 & 0.0786 &  0.3945  &   88.74  &  95.84  &  98.59  \\
     \cline{2-10}&  & OmniFusion & CVPR'22 & 0.1154 & 0.0775 & 0.3809 & 86.74 & 96.03 & 98.71 \\
     S2D3D& & UniFuse & RAL'21  & 0.1124 &  0.0709  & 0.3555 & 87.06 & 97.04 &  98.99 \\
     & ResNet-34& Elite360D & CVPR'24  & 0.1182 & 0.0728 & 0.3756 & 88.72 & 96.84 & 98.92 \\
     \cline{3-10}& & Ours  & - & \textbf{0.1029} & \textbf{0.0599} & \textbf{0.3228} & \textbf{89.13} & \textbf{96.87} & \textbf{99.06}\\
    \hline  & Transformer & EGFormer &  ICCV'23   & 0.2205 &  0.4509  & 0.6841 &  79.79  &  90.71  & 94.55 \\
     & & PanoFormer &  ECCV'22   & 0.2549 & 0.4949 & 0.7937 & 74.70 & 89.15 & 93.97 \\
     \cline{2-10}
     Struct3D& & BiFuse &  CVPR' 20    & 0.1573 & 0.2455 & 0.5213 &   85.91  & 94.00 & 96.72 \\
     & & UniFuse & RAL' 21 & 0.1506 & 0.2319 & 0.5016 & 85.42 & 93.99 & 96.76 \\
     & ResNet-34& Elite360D &CVPR'24  & 0.1480 &\textbf{0.2215} & 0.4961 & 87.41 & 94.34 & 96.66 \\
     \cline{3-10}& & Ours &- & \textbf{0.1401} & 0.2499 & \textbf{0.4394} & \textbf{87.78} & \textbf{94.49} & \textbf{96.82} \\
    \hline
    \label{tab:quanti}
    \end{tabular}
    \vspace{-10pt}
    \begin{tablenotes}
     \item The bold values indicate the best performance.
   \end{tablenotes}
  \end{threeparttable}
\end{table*}

\subsection{Loss Function} We propose an objective function that combines reverse huber loss and spherical gradient loss to jointly optimize pixel-wise depth accuracy and spatial gradient consistency of the predicted depth maps. The Reverse Huber Loss is defined as:

\begin{eqnarray}
 \mathcal{L}_\beta(D, G) = \frac{1}{N} \sum_{i \in N} 
\begin{cases}
|D_i - G_i| & \text{if } |D_i - G_i| \leq \theta, \\
\frac{(D_i - G_i)^2 + \delta^2}{2\delta} & \text{otherwise},
\end{cases}   
\end{eqnarray}
where \(D_i\) and \(G_i\) are the predicted and ground truth depth values at the \(i\)-th pixel, respectively, and \(N\) is the total number of pixels in the depth map. The threshold \(\theta = 0.2\) determines when the error is treated linearly or quadratically, while \(\delta\) is a small constant that smooths the quadratic penalty for large errors. This loss function applies a linear penalty for small errors and a smooth quadratic penalty for large errors, providing robustness against noise and outliers.

The spherical gradient loss is defined as:
\begin{eqnarray}
 \mathcal{L}_g = |G_x(G) - G_x(D)| + |G_y(G) - G_y(D)|,
\end{eqnarray}
where \(G\) and \(D\) represent the ground truth and predicted depth maps, respectively, and \(G_x\) and \(G_y\) denote the gradients of these depth maps along the horizontal and vertical directions. This loss encourages the predicted depth map to maintain spatial gradient consistency with the ground truth, helping preserve smoothness and structure.

Finally, the overall objective function is:

\begin{eqnarray}
\mathcal{L} = \mathcal{L}_\beta(D, G) + \eta \mathcal{L}_g,    
\end{eqnarray}
where \(\eta = 0.5\) is a hyperparameter that balances the contributions of the two loss terms. By jointly optimizing these two losses, the model learns to predict accurate depth values while maintaining spatial structure and smoothness in the predicted depth maps.

\begin{table}[t]
\centering
\caption{Complexity comparison with the SOTA methods.}
\begin{tabular}{c c c c}
\hline Backbone & Method & Params (M) & FLOPs   \\
\hline 
Transformer & EGFormer  & 15.39 & 66.21 \\
 & PanoFormer  & 20.38 & 81.09 \\
\cline{1-4}  & BiFuse  & 56.01 &  199.58\\
  & BiFuse++ & 52.49 & 87.48 \\
 ResNet-34 & UniFuse & 50.48 & 96.52\\
 & OmniFusion & 42.46 & 142.29\\
 & HRDFuse  &  46.31 & 80.87\\
 & Elite360D &  25.54 & 65.29\\
 & Ours &30.75 & 75.67 \\
\hline
\end{tabular}
\label{tab:Complex}
\end{table}
\section{Experiment}
\label{section4}
To validate the effectiveness of the proposed scheme, we conduct experiments on three popular datasets. The basic experiment settings are as follows.

We conduct all experiments on a single NVIDIA GeForce RTX 3090 GPU, implementing our approach using the PyTorch framework. We use the Adam optimizer \cite{kingma2014adam} with an initialized learning rate of $1\times 10^{-4}$ to train our model. We keep the default settings for other optimizer hyperparameters. During the training period, we use a batch size of 8 and employ left-right flipping, panoramic horizontal rotation, and luminance change for data augmentation, as previously proposed in works \cite{Jiang2021UniFuseUF, Shen2022PanoFormerPT}. The backbones we employed are all pre-trained on ImageNet-1K. 

\noindent \textbf{Dataset.} Two real-world datasets for experimental validation: Stanford 2D-3D~\cite{Armeni2017Joint2D}, and Matterport3D~\cite{chang2017matterport3d}. The depth maps in Stanford 2D-3D are constructed from reconstructed 3D models, and some depths are inaccurate. Besides, the bottom and top regions of the RGB images are missing. We follow the official splits and get 1,040 RGB-D pairs for training and 343 for testing. The Matterport3D dataset provides raw depth whose top and bottom depths are missing. For the RGB panoramas in Matterport3D, we follow the Unifuse method \cite{jiang2021unifuse} to fill the missing regions. Finally, we get 7829 pairs for training and 2014 pairs for testing. 

One virtue dataset: Structured3D~\cite {Zheng2019Structured3DAL}. The virtue dataset was first proposed to explore the layout estimation task. Recently, some researchers have preferred to evaluate the depth estimation models on the dataset for its high-quality annotations. We follow the official splits that we employ the frame scene\_0000 to frame scene\_02999 for training and scene\_03250 to scene\_03499 for testing. We especially follow the official guidelines and remove invalid cases.

For all the datasets, we adopt a resolution of 1024×512 (width×height, 3 channels) of the single input panorama.
\begin{figure*}[t]
  \centering
  \includegraphics[width=\textwidth]{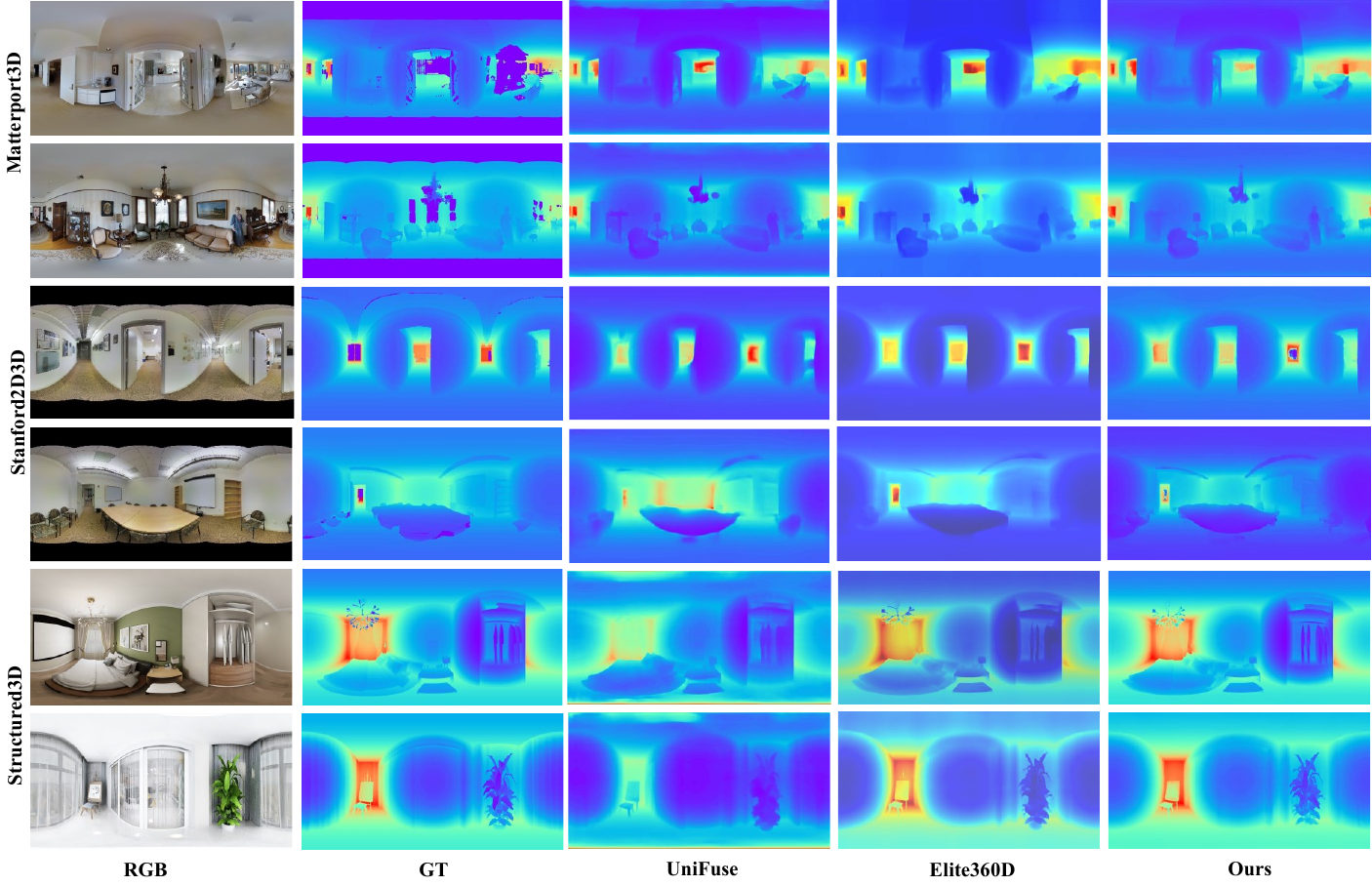} 
  \caption{Qualitative comparison results. We exhibit the predicted depth maps of the Unifuse~\cite{Jiang2021UniFuseUF}, Elite360D~\cite{Ai_2024_CVPR}, and ours. All depth maps adopt a unified visualization method.} 
  \label{fig:qualitative}
\end{figure*}

\begin{figure*}[t]
  \centering
  \includegraphics[width=\textwidth]{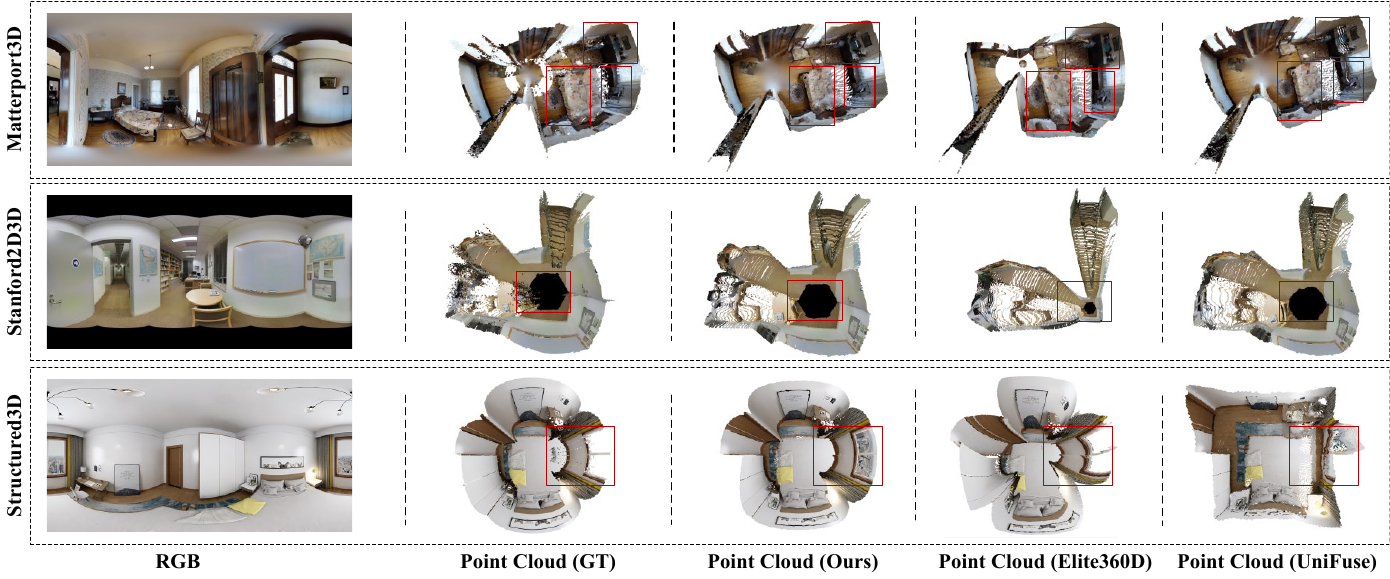} 
  \caption{Visualization of the point cloud generated with the ground truth (GT) and the predicted depth map (Ours, Elite360D and UniFuse). The areas marked in red rectangles indicate that our approach gains superior performance (dense and high-quality point cloud).} 
  \label{fig:ptc}
\end{figure*}
\subsection{Evaluation Metrics}
For evaluation, we follow the previous works \cite{Jiang2021UniFuseUF, Shen2022PanoFormerPT,Yun2023EGformerEG,Ai_2024_CVPR} and employ the standard evaluation metrics, including absolute relative error (Abs Rel), squared relative error (Sq Rel), Root Mean Squared Error (RMSE), and $\delta_{t}, t \in \{1.25, 1.25^2, 1.25^3\}$. Among them, $\delta_{t}$ provides a better reflection of the accuracy of the value per depth. RMSE indicates the average difference between the predicted and the actual value. Following Elite360D \cite{Ai_2024_CVPR}, we also report the number of parameters and FLOPS to evaluate the efficiency of our approach.

\subsection{Comparison Analysis}
\textbf{Quantitative Results.} We exhibit the quantitative comparison results in Table \ref{tab:quanti}, which demonstrate that our model outperforms all other state-of-the-art solutions on most metrics. Our PGFuse method significantly outperforms the compared methods. Specifically, on the real-world Matterport3D and Stanford2D3D datasets, our approach demonstrates notable improvements in the important RMSE and $\delta_1$ metrics. Additionally, it achieves comparable results on the virtual Structured3D dataset, showcasing its robustness and effectiveness across various environments. These results highlight the efficacy of PGFuse in monocular 360 depth estimation across diverse and challenging datasets. For the Matterport3D dataset, PGFuse consistently outperforms other methods on the accuracy metrics with stricter thresholds $\delta_1$ and $\delta_2$, as well as on RMSE and Abs Rel. However, on the metric $\delta_3$, PGFuse exhibits slightly lower performance than that of Elite360D~\cite{Ai_2024_CVPR} and UniFuse~\cite{jiang2021unifuse}. Since the $\delta_3$ metric uses the loosest threshold and is almost saturated for near- and mid-range depths for all methods in comparison, it is relatively insensitive to the improvements brought by our design, which mainly targets local details. Note that in the small-scale, realistic Stanford2D3D dataset, PGFuse outperforms other methods in all metrics. In \cite{Ai_2024_CVPR}, the authors argue that due to the limited data in the Stanford2D3D dataset, the transformer-based approach can hardly provide accurate global perception, while the core components of our method are CNN-based and cannot be influenced by this negative effect. As for the Structured3D dataset, our method achieves superior performance on most metrics, but the Sq Rel value is noticeably higher than that of UniFuse and Elite360D. We attribute this primarily to the depth preprocessing protocol. Specifically, for a fair comparison, we follow Elite360D by normalizing the depth maps and rescaling them to a maximum range of 10\,m. This global rescaling changes the original depth distribution and introduces a slight shift in depth scale. Since PanoGabor estimates depth from Gabor-based learning of local texture patterns, a scale mismatch between the normalized depths and the original scene geometry can disturb this texture–depth mapping. Under this preprocessing protocol, such shifts in the depth scale tend to produce a modest increase in the Sq Rel metric. 

 \textbf{Qualitative Results.} We exhibit the qualitative comparison results with the current SoTA UniFuse \cite{Jiang2021UniFuseUF} and Elite360D \cite{Ai_2024_CVPR} method. As shown in Fig. \ref{fig:qualitative}, we can observe that UniFuse can effectively remove the negative effect of the panoramic distortions, but it still suffers from losing detailed information. This degradation is due to the projection format of the cubemap that loses 25\% of pixels when converted to ERPs. In contrast, Elite360D alleviates such information loss and provides relatively sharper depth boundaries, yet it fails to recover fine-grained structural cues in complex scenarios. Our method preserves a competitive depth range and maintains accurate details. This preservation of details contributes to superior performance in RMSE metrics. In addition, from Fig. \ref{fig:ptc} our method shows remarkable results even in the regions where the ground truth is wrong.

\textbf{Complexity.} We show the complexity comparison results in Table \ref{tab:Complex}. We can observe that those transformer-based architectures may have fewer parameters and FLOPs. However, the complexity of these approaches is quadratic to the input pixels, which limits their application to high-resolution images. In addition, their graphics memory cost is huge, which makes them difficult to use in practice. The other SoTA approach, Bifuse, has the largest parameters and huge FLOPs, and the training time is unbearable for large-scale datasets. Although our scheme has slightly more parameters and FLOPs than Elite360D, the proposed approach performs better than Elite360D, especially on the RMSE metric, by a significant margin. In addition, we argue that the core fusion component of Elite360D is attention-based, similar to transformer-based frameworks, resulting in fewer parameters and FLOPs.  To further evaluate computational efficiency, we benchmark both inference speed and memory on a single NVIDIA RTX 3090 (24 GB) under the same setting (batch size = 8). Our method achieves 40 FPS with 5770 MB GPU memory, whereas Elite360D attains 21 FPS with 7728 MB. This corresponds to 25 ms/image for our method versus 48 ms/image for Elite360D, demonstrating balanced efficiency by delivering both higher throughput and lower memory consumption under identical hardware conditions.

\begin{table*}[h]
\setlength{\belowcaptionskip}{-20pt}
\begin{threeparttable}
\caption{Ablation study.}
\centering
    \begin{tabular}{c c c c c c c c c c c c}
    \hline Datasets &Index& Fusion Module & Gabor Type & Spherical Gradient  &$\mathrm{RMSE} \downarrow$   & Sq Rel  $\downarrow$  & Abs Rel  $\downarrow$   &  $\delta_{1}(\%) \uparrow$  &  $\delta_{2}(\%) \uparrow$  &  $\delta_{3}(\%) \uparrow$  \\
    \hline 
    &a& Cat & \ding{55}   & \ding{55}  & 0.5021 & 0.0981 & 0.1125 &  87.29  & 95.72 & 98.19 \\
    &b& CS-UFM &\ding{55}     &\ding{55}   & 0.4906 &  0.0971  & 0.1107 &   88.06  & 96.09 & 98.39 \\
    M3D&c& CS-UFM & PanoGabor   &\ding{55}   & 0.4767 &   0.0903  &  0.1086  & 88.51 & 96.34 & 98.54 \\
     &d& CS-UFM & PanoGabor   &\checkmark  & \textbf{0.4624} & \textbf{0.0858} & \textbf{0.1067} & \textbf{89.06} & \textbf{96.75} & \textbf{98.72} \\
     \cline{2-11}
     &e& CS-UFM & Normal Gabor   &\checkmark  &0.4848  & 0.0917 & 0.1089 & 88.25 & 96.25 & 98.46\\
    \hline
    \label{tab:ablation}
    \end{tabular}
    \vspace{-10pt}
    \begin{tablenotes}
     \item The bold values indicate the best performance.
   \end{tablenotes}
  \end{threeparttable}
\end{table*}

\subsection{Ablation Study}
To demonstrate the effectiveness of the main proposed components, we follow Unifuse \cite{Jiang2021UniFuseUF} and Elite360D \cite{Ai_2024_CVPR} to design the ablation study on the Matterport3D dataset \cite{Chang2017Matterport3DLF}. All the models have the same conditions and are trained to converge. As illustrated in Fig. \ref{fig:framework}, we remove the fusion stage from the full framework to serve as the baseline for conducting the ablation study. 
The experimental results of the baseline are presented as `a' in Table \ref{tab:ablation}. We further add the proposed CS-UFM module, a version without the PanoGabor layer embedded as `b'. Index `c' refers to the use of the full CS-UFM module. Index `d' indicates that we employ the proposed spherical gradient constraint to stabilize the orientation sensitivity of the Gabor filters. Additionally, in `e', all PanoGabor layers are replaced with normal Gabor filters to further verify the effectiveness of the designed PanoGabor.

\begin{figure}[t]
  \centering
  \includegraphics[width=0.48\textwidth]{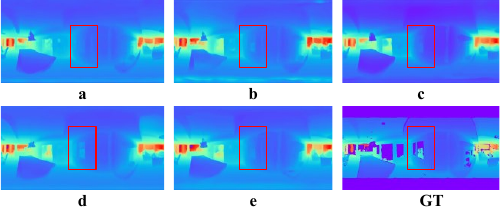} 
  \caption{Qualitative comparison results of ablation study. The areas marked in red rectangles indicate that our approach gains superior performance.} 
  \label{fig:abla}
\end{figure}

\begin{figure}[h]
  \centering
  \includegraphics[width=0.48\textwidth]{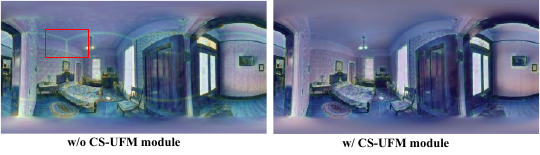} 
  \caption{Visualization of the heatmap of the features w/o and w/ the CS-UFM module. Because each face in the cubemap projection is processed individually, resulting in discontinuous ERP features. The CS-UFM module can eliminate this issue effectively.} 
  \label{fig:fus}
\end{figure}

\textbf{Effectiveness of the CS-UFM module.} From Table \ref{tab:ablation} and Fig. \ref{fig:abla}, we can observe that the CS-UFM module can obtain an improvement even without the designed PanoGabor layer. This is because the fusion module can effectively fuse the two sets of features due to the channel-wise and spatial-wise designs. Moreover, when further embedding the proposed PanoGabor layer, our approach gets a significant improvement across all the evaluation metrics, verifying the necessity of addressing the reintroduced distortion problems during the fusion process. From Fig. \ref{fig:fus}, we can observe that our CS-UFM module has a uniformly distributed heat map, indicating that it can effectively fuse features and handle the reintroduced distortion.

\textbf{Effectiveness of the PanoGabor.} Comparing b and e in Table \ref{tab:ablation} and Fig. \ref{fig:abla}, we can observe that, benefiting from the frequency domain transformation along with the capability of reducing surface texture interference, the normal Gabor layer can benefit the fusion process. From d and e, we can observe that our designed PanoGabor performs better than the normal ones due to the distortion-aware designs. 
\begin{figure}[h]
    \centering
   \includegraphics[width=0.4\textwidth]{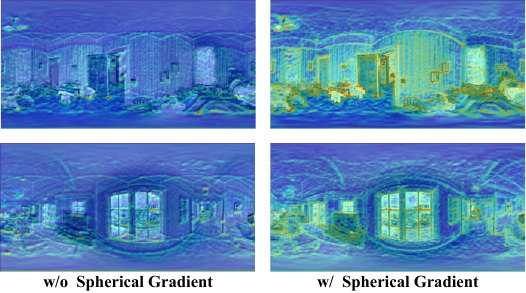}
    \caption{Qualitative comparison of feature responses with and without the Spherical Gradient constraint.}
    \label{fig:sg}
\end{figure}

\textbf{Effectiveness of the spherical gradient constraint.} Recently, some works \cite{Shen2022PanoFormerPT, Ai_2024_CVPR} prefer to employ a gradient-based constraint to sharpen the depth edges. This strategy aims to extract the horizontal/vertical structures to design the objective function. However, since the Gabor filters are orientation-sensitive, we design the spherical gradient constraint to incorporate the panoramic geometry to stabilize this orientation sensitivity.

 From Table \ref{tab:ablation}, we observe that the PanoGabor with the proposed spherical gradient constraint outperforms the version without it (i.e., settings (c) vs. (d) in Table~\ref{tab:ablation}). Removing the constraint consistently degrades all evaluation metrics (e.g., RMSE rises from 0.4624 to 0.4767, $\delta_1$ drops from 89.06 to 88.51). Importantly, this degradation is consistent across both error- and accuracy-based measures, suggesting that orientation sensitivity introduces instability at both global and local levels.

As shown in Fig.~\ref{fig:sg}, the feature responses without the constraint exhibit pronounced orientation-dependent variations: edges aligned with certain directions are overly amplified, while others are suppressed or distorted, resulting in visible artifacts such as uneven contrast and directionally biased activations. In contrast, applying the constraint produces smoother and more homogeneous responses across orientations, where structural boundaries are preserved more uniformly and spurious artifacts are greatly reduced. This visual evidence reveals the underlying cause of the observed performance drop and demonstrates that the spherical gradient constraint effectively enforces orientation-robust feature representations.
\begin{figure*}[t]
    \centering
   \includegraphics[width=\textwidth]{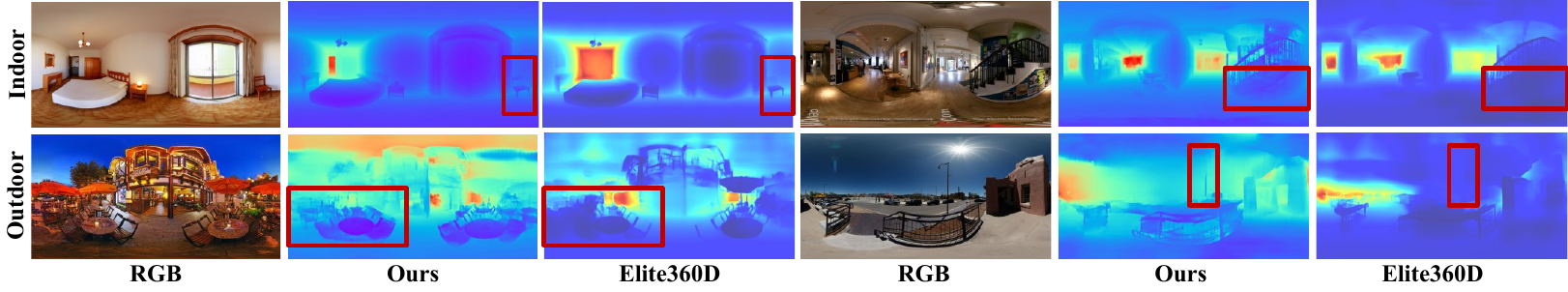}
    \caption{Qualitative comparison results on indoor (top) and outdoor (bottom) scenes.}
    \label{fig::outdoor}
\end{figure*}

 \textbf{Generalization Analysis.}
We evaluate the generalization ability of PGFuse on the SUN360 dataset~\cite{xiao2012recognizing}, which contains diverse panoramic images without depth annotations. Since ground truth is unavailable, we report qualitative results in Fig.~\ref{fig::outdoor}.

As shown in the outdoor examples, PGFuse yields plausible estimates of near-range structures such as buildings and ground surfaces, while predictions for distant regions are less precise. This discrepancy arises from the inherent differences between indoor and outdoor data: (1) indoor training datasets constrain the depth range to within 10 m, whereas outdoor scenes span substantially larger ranges; and (2) outdoor panoramas often include large homogeneous regions (e.g., sky) and more diverse illumination conditions. This illustrates the intrinsic challenge of transferring a model trained on indoor data (with depth limited to 10 m) to outdoor environments.

As shown in Fig.~\ref{fig::outdoor}, PGFuse provides consistent and structurally reliable depth maps, with well-preserved object boundaries and relative depth relations. In comparison, Elite360D\cite{Ai_2024_CVPR} also produces reasonable results, but tends to lose details in complex regions (highlighted in red boxes), particularly under outdoor settings. These findings confirm that PGFuse generalizes robustly within the indoor domain, while still delivering plausible outdoor predictions despite the increased difficulty.

\begin{table}[h]
    \centering
    \caption{Sensitivity of model performance to Gabor parameter initialization.
    }
            \begin{tabular}{c@{\hspace{10pt}} c@{\hspace{10pt}} c@{\hspace{10pt}} c@{\hspace{10pt}} c@{\hspace{10pt}}}
    \hline 
    Init. Gabor Param. &$\mathrm{RMSE} \downarrow$   & Sq Rel  $\downarrow$  & Abs Rel  $\downarrow$   &  $\delta_{1}(\%) \uparrow$\\  
    \hline 
    Random& \multicolumn{4}{c}{Non-convergence} \\
    Xavier(U)& 0.4856 &  0.0933  & 0.1127 &   87.34\\
    Gabor Init.&0.4848  & 0.0917 & 0.1089 & 88.25\\
    Ours& \textbf{0.4624} & \textbf{0.0858} & \textbf{0.1067} & \textbf{89.06}\\
    \hline
\label{tab:sens}
\end{tabular}
\vspace{-10pt}
\begin{tablenotes}
\item The bold values indicate the best performance.
\end{tablenotes}
\end{table}

 \textbf{Analysis of initialization sensitivity for Gabor parameters.} In Table~\ref{tab:sens}, we evaluate four initialization strategies for Gabor parameters: random initialization leads to non-convergence; Xavier initialization~\cite{glorot10a} converges but gives suboptimal accuracy since it ignores the structured nature of Gabor filters; handcrafted Gabor initialization~\cite{alekseev2019gabornet} provides frequency-selective priors but neglects panoramic distortions; and our distortion-aware initialization achieves the best performance by aligning orientations and frequencies with spherical geometry. These results confirm that PGFuse is sensitive to initialization, but the proposed distortion-aware scheme can effectively alleviate this sensitivity, stabilize training, and improve accuracy.

\begin{table}[h]
\caption{Ablation study on various backbones.}
\centering
\begin{tabular}{c@{\hspace{10pt}} c@{\hspace{10pt}} c@{\hspace{10pt}} c@{\hspace{10pt}} c@{\hspace{10pt}} c@{\hspace{10pt}} c@{\hspace{10pt}}}
\hline Dataset& Backbone & Abs Rel  $\downarrow$  & Sq Rel  $\downarrow$  &  $\mathrm{RMSE} \downarrow$  &  $\delta_{1}(\%) \uparrow$\\
\hline 
  \multirow{3}*{M3D}&ResNet-18
    &0.1125& 0.0977& 0.4933& 87.78\\

    &ResNet-34
    & 0.1067& 0.0858& 0.4624& 89.06\\

    &ResNet-50 
 &0.1058& 0.0841& 0.4618& 88.93\\
    \hline
\end{tabular}
\begin{tablenotes}
\item The bold values indicate the best performance.
\end{tablenotes}
\label{tab:backbone}
\end{table}

\textbf{Various Backbones.} Following Elite360D \cite{Ai_2024_CVPR}, we also adopt different backbones for our framework. From Table \ref{tab:backbone}, we can observe that our scheme can benefit from employing more powerful networks. In particular, PGFuse demonstrates strong scalability and can effectively leverage stronger backbones. Unlike UniFuse, which suffers from limited performance gains due to reintroduced distortion when upgrading the backbone\cite{jiang2021unifuse}, PGFuse maintains consistent improvements across ResNet-18, ResNet-34, and ResNet-50, clearly validating the robustness of our design against structural distortion.
\begin{table}[h]
\caption{More advanced performance comparison results. * represent that Elite360D employs Efficient-B5 as backbone. $^\star$ indicates that we replace one of the cubemap features with extracted ERP features.}
\centering
\resizebox{0.5\textwidth}{!}{
\begin{threeparttable}[b]
\begin{tabular}{c c c c c c c c}
\hline Datasets &Method& Abs Rel  $\downarrow$  & Sq Rel  $\downarrow$  &  $\mathrm{RMSE} \downarrow$  &  $\delta_{1}(\%) \uparrow$\\
\hline 
M3D&Elite360D*& 0.1048 & 0.0805 & 0.4524 & 89.92\\
M3D&Ours$^\star$ &\textbf{0.0972} &\textbf{0.0801} &\textbf{0.4480} & \textbf{90.67}  \\
\hline
\end{tabular}
\begin{tablenotes}
\item The bold values indicate the best performance.
\end{tablenotes}
\end{threeparttable}}
\label{tab:trade-off}
\end{table}

\textbf{Trade-off between performance and efficiency.} From Table \ref{tab:trade-off}, Elite360D has obtained a significant improvement by employing EfficientNet-B5 as a feature extractor. However, it brings unbearable memory costs. In contrast, our approach can achieve superior performance just by replacing a group of input features without additional memory cost. This also demonstrates the robustness and scalability of our PGFuse.

\textbf{Discussion of Projection Trade-offs.} Although PGFuse is formulated as a two-stage distortion-optimization pipeline to address different distortion problems in complementary domains, the same pipeline can be instantiated with alternative projection choices (e.g., ERP-only or ERP+cubemap), providing practical flexibility to balance global consistency, local detail accuracy, and computational cost according to specific application requirements.

Specifically, we evaluate three settings: ERP-only, dual-cubemap (our default configuration), and ERP+cubemap. The ERP-only setting estimates panoramic depth directly in the full ERP domain. It provides a globally consistent representation and a lower computational cost. However, PanoGabor would have to be incorporated into the backbone to explicitly handle panoramic distortion. This modification changes the architecture on which the weights were pretrained and typically weakens the effect of pretraining; thus, this setting struggles to capture local details. In contrast, the dual-cubemap setting adopts a two-stage distortion-optimization pipeline. It first extracts features in the low-distortion cubemap domain and then refines them in the ERP domain with PanoGabor. This design largely reduces the negative effects of panoramic distortion while keeping the computational cost acceptable. It produces finer local details and comparable global consistency. The ERP+cubemap setting combines the strengths of the two settings above. However, it requires two separate backbones to extract features from the ERP and cubemap representations, which leads to significantly higher computational cost.
\section{Conclusion}
In this paper, we have proposed a Gabor-based framework, termed PGFuse, for 360 depth estimation. Since the existing fuse-based scheme ignores the reintroduced distortion when integrating other representations into ERP format, our proposed PGFuse embeds a Gabor-based layer into a unidirectional fusion module to explicitly deal with the reintroduced distortion. Specifically, to make the Gabor filters adapt to 360 images, we first introduce a latitude-aware representation to model the panoramic distortion distribution pattern. Then we leverage this representation to generate customized Gabor filters. These tuned PanoGabor can enhance the depth cues (\textit{i.e.}, resistant to surface texture and lighting interference, while preserving structural details) in the frequency domain and achieve effective distortion awareness. Therefore, our fusion module can deliver spatial-continuous and content-consistent ERP features. Extensive experiment results demonstrate that our approach outperforms current SoTA approaches.

{\small
\bibliographystyle{IEEEtran}
\bibliography{reference}
}

\end{document}